\documentclass[10pt,journal,compsoc]{IEEEtran}

\usepackage[T1]{fontenc}
\usepackage[utf8]{inputenc}
\usepackage{cite}
\usepackage{graphicx}
\usepackage{booktabs}
\usepackage{array}
\usepackage{url}
\usepackage{float}
\usepackage{comment}

\usepackage[table]{xcolor}   
\usepackage{array}           
\usepackage{amssymb}         

\definecolor{linkblue}{HTML}{1F5F99}
\definecolor{citegray}{HTML}{4A4A4A}

\definecolor{tierNative}{HTML}{D6EBD3}  
\definecolor{tierAug}{HTML}{D4E4F7}     
\definecolor{tierBound}{HTML}{FBE7CF}   
\definecolor{tierOut}{HTML}{ECECEC}     
\definecolor{headDark}{HTML}{1F4E79}    
\definecolor{headLite}{HTML}{D9EAF7}    
\usepackage{xcolor}
\definecolor{linkBlue}{HTML}{1F4E79}
\usepackage[
  colorlinks=true,
  linkcolor=black,
  citecolor=black,
  urlcolor=linkBlue
]{hyperref}

\begin{document}

\title{AI-Native Games: A Survey and Roadmap}

\newcommand{\affmark}[1]{\textsuperscript{#1}}
\author{
\IEEEauthorblockN{
Zhiyue Xu \affmark{1,2},
Fandi Meng \affmark{2,3},
Kaijie Xu \affmark{4},
Clark Verbrugge \affmark{4},
Simon Lucas \affmark{5},
Jian Zhao \affmark{2,3}
}\\
\IEEEauthorblockA{
\affmark{1}Institute of Automation, Chinese Academy of Sciences \quad
\affmark{2}Zhongguancun Academy\\
\affmark{3}Zhongguancun Institute of Artificial Intelligence \\
\affmark{4}School of Computer Science, McGill University, Montreal, Quebec, Canada\\
\affmark{5}Game AI Research Group, Queen Mary University of London, London, United Kingdom\\
Email: xuzhiyue2026@ia.ac.cn, mengfandi@zgci.ac.cn, kaijie.xu2@mail.mcgill.ca, clump@cs.mcgill.ca, simon.lucas@qmul.ac.uk, jianzhao@zgci.ac.cn\\
}}

\maketitle

\begin{abstract}

Generative AI now enables games to produce dialogue, quests, characters, images, and worlds at runtime. Yet generation alone does not make a game AI-native, nor does it guarantee playability. This paper defines AI-native games by whether runtime generative AI is constitutive of the core loop: if the AI component were removed or trivially replaced, the central form of play would collapse or become fundamentally different. This counterfactual criterion separates AI-native games from AI-augmented games and adjacent boundary artifacts. Using this definition, we screen candidate artifacts and analyze 53 publicly available AI-native games and prototypes. We introduce a dual-axis G/N taxonomy: the G-axis captures player-facing game type, while the N-axis captures the dominant AI mechanic that makes generative AI indispensable to play. The corpus is concentrated around language-forward designs, especially narrative adventure, epistemic interaction, and generative narrative, while categories such as semantic adjudication, multi-agent simulation, generative construction, and relationship/companion play remain less represented.
We argue that the central design problem is organizing semantic openness into stable gameplay. AI-native design depends on mechanical invariants: goals, rules, state, feedback, pacing, and player agency that make open-ended AI outputs interpretable and consequential. We conclude with a roadmap for controllable generation, AI-as-mechanic design, multimodal and multi-agent systems, inference economics, evaluation, safety, and regulation. 
The curated game corpus, source links, and coding records are available at \href{https://github.com/igloomatics/Awesome-AI-native-games-collection}{Awesome-AI-native-games-collection} repository.
\end{abstract}


\IEEEpeerreviewmaketitle

\section{Introduction}
\label{sec:introduction}

Generative artificial intelligence is increasingly moving from a production-side tool for game development to a runtime component of game systems. Earlier
computational and generative methods were mostly applied before play, helping designers, artists, writers, and programmers create or explore assets, levels, rules, and code \cite{shaker2016pcg}. A more consequential shift for game design is now emerging: foundation models are being integrated into games during play, where they interpret player input, generate responses, update fictional worlds, control characters, and sometimes arbitrate outcomes. AI is no longer only a tool used to make the game; it becomes part of the playable system itself \cite{gallotta2024llmGames,yang2024gptGames,maleki2024pcg,sun2023languageReality}.


Runtime generation, however, does not by itself produce engaging gameplay. Games depend on constraints, goals, feedback, and opportunities for strategic understanding \cite{hunicke2004mda,sweetser2005gameflow}. Generative models invoked at runtime may produce plausible dialogue, quests, items, or worlds, but these outputs can be inconsistent, unbalanced, unsafe, or disconnected from player agency. Unbounded generation can therefore increase variety while weakening the very structures that make play meaningful. 
The question, then, is how generative AI can move beyond content generation and become part of reliable, compelling game mechanics\cite{gallotta2024llmGames,sweetser2024llmVideoGames}.

To study this question, we adopt the notion of \emph{AI-native games} articulated by Sun et al.\ in their work on \emph{1001 Nights} \cite{sun2023languageReality}. It is also informed by recent industry discussions that similarly emphasize the indispensability of AI to a game's core mechanic \cite{mason2026stateAINativeGames}. Following this view, we define an AI-native game as one in which generative AI functions as a constitutive mechanism of the core loop: were the AI component removed, the core gameplay could no longer operate as designed, or would become a fundamentally different form of play. This definition does not require every asset, rule, or interaction to be generated by AI. It requires only that generative AI be present during play, directly involved in the core loop, and not trivially replaceable by finite authored assets or conventional deterministic mechanisms.

This distinction matters because digital games have long employed AI in powerful forms that do not, by themselves, make a game AI-native.
Traditional game AI (e.g., pathfinding, finite-state machines, behavior trees, utility systems, adaptive difficulty, opponent modelling) can produce complex and responsive behaviour, but it operates over goals, states, actions, and evaluation criteria fixed during development \cite{millington2009artificial, hunicke2005dda,yannakakis2018aigames}. Such AI supports the functioning of the game world without expanding what the player can meaningfully say, attempt, or transform at runtime.

Two earlier traditions nonetheless point toward a broader role for AI in play. The first is interactive drama, exemplified by Mateas and Stern's \emph{Fa{\c{c}}ade}. The game used natural language understanding, believable agents, beat sequencing, and drama management to make social interaction and narrative tension central to gameplay \cite{mateas2005structuring,mateas2002abl,riedl2013interactiveNarrative}. Its significance lies less in anticipating modern generative models than in demonstrating that player language, character response, and dramatic progression could serve as the primary interactive material of a game. The second is procedural content generation (PCG) and related runtime adaptive systems. Offline PCG showed that algorithms can synthesize levels, items, quests, textures, and other artifacts \cite{togelius2011searchbased,hendrikx2013pcg,shaker2016pcg,summerville2018proceduralcontentgenerationmachine}, while adaptive systems such as the \emph{Left 4 Dead} ``AI Director'' further demonstrated that pacing, enemy spawns, and resource distribution can be modulated in response to player state during play \cite{yannakakis2011experience,hunicke2005dda,valve2009aidir}. Both traditions suggested, well before large language models (LLMs), that parts of content generation and experience regulation could be deferred to runtime.

However, these predecessors remain bounded. Adaptive PCG systems may evaluate, search, or optimize in real time, but they do so within authored representations, parameter ranges, objective functions, or content libraries. In other words, they select and assemble within a bounded design space, but they do not usually interpret open-ended player input or generate new game meanings. AI-native games mark a further step by using generative foundation models to interpret language, generate context-sensitive content, maintain conversational and narrative state, enact social roles, and transform text, image, audio, or code-like representations during play \cite{gallotta2024llmGames,sweetser2024llmVideoGames,yang2024gptGames}.

The resulting landscape is conceptually blurred: many contemporary games incorporate generative AI, but only some make it constitutive of gameplay. Surveying AI-native games therefore requires distinguishing them from adjacent forms of AI use, constructing an explicit corpus, and developing a classification that captures both the player-facing form of a game and the AI mechanism that organizes its play.

This survey makes four contributions. 
First, it defines AI-native games through runtime generative AI, core-loop dependence, and non-substitutability, and introduces a 2$\times$2 boundary matrix that separates AI-native games, AI-augmented games, AI-boundary artifacts, and out-of-scope cases. 
Second, it constructs and analyzes a corpus of 53 publicly available AI-native games and prototypes selected from candidate artifacts. 
Third, it proposes a dual-axis G/N classification framework, where game type (G) captures the player-facing form and dominant AI mechanic (N) captures the AI function that makes generative AI constitutive of play. 
Fourth, it uses this framework to identify current design concentrations and gaps, and develops a roadmap from rule-bounded generation and mechanics-level design to runtime inference, model dependence, evaluation, safety, and regulation.

The remainder of this article proceeds as follows. Section~\ref{sec:methodology} presents the conceptual starting point, review scope, corpus construction, and coding procedure. Section~\ref{sec:definition} defines AI-native games and situates them within the historical and contemporary landscape of game AI, PCG, interactive drama, and runtime generative play. Section~\ref{sec:classification} develops an inductive classification of AI-native games and reports descriptive statistics from the analyzed corpus. Section~\ref{sec:roadmap} presents a roadmap for future research. Section~\ref{sec:ethics} discusses ethics, safety, and regulatory implications. Section~\ref{sec:conclusion} concludes by summarizing the argument and identifying the broader significance of AI-native games.

\section{Review Scope and Methodology}
\label{sec:methodology}


The definition introduced in Section~\ref{sec:introduction} grows out of several earlier lines of AI-centered game research, including Expressive AI, interactive drama, believable agents, drama management, and AI-driven interactive narrative~\cite{mateas2001expressiveAI,mateas2003facade,mateas2002abl,riedl2010narrativePlanning}. These works showed, well before the current wave of foundation models, that AI could serve as playable material by shaping social interaction, narrative response, and character behavior. Recent surveys on LLMs in games, GPT-based game research, LLM-assisted development, LLM-based procedural content generation, game agents, mixed-initiative design, and broader game AI have mapped many of the surrounding techniques and applications~\cite{gallotta2024llmGames,sweetser2024llmVideoGames,yang2024gptGames,yang2025gptGames,maleki2024pcg}. However, they usually take the model, method, or application area as the main unit of analysis. This leaves a more design-specific question less directly addressed: when does runtime generative AI become part of the game itself, rather than a tool, feature, or content source around it? We use this question to frame AI-native games as a gameplay category, and focus on how such games can be defined, delimited, classified, and evaluated.

For the empirical analysis, we use a qualitative artifact-analysis approach and focus on recent post-LLM games and public prototypes. After screening, the main corpus contains 53 AI-native artifacts; \emph{Fa{\c{c}}ade} is discussed only as a historical precursor. We also examined nearby cases that did not meet the inclusion criteria, such as production-only AI uses, static AI-generated assets, optional chat NPCs, chatbot or tavern-style role-play systems, concept-only demonstrations, and tools without verifiable game structure.

Because the taxonomy in Section~\ref{sec:classification} was developed through inductive coding rather than imposed from a predefined genre list, we first constructed a broad candidate pool of AI-native and near-boundary artifacts. The candidate search focused on the post-LLM period, especially artifacts publicly available from 2019 onward, when LLM-enabled interaction made free-form input, narrative continuation, flexible NPC response, and semantic action interpretation more widely practicable. Candidate artifacts were collected through two complementary routes. First, academic prototypes, historical precedents, and research cases were identified through academic references and related literature on AI-native games, LLM-driven games, LLMs in and for games, PCG, interactive narrative, AI agents, and game AI~\cite{gallotta2024llmGames,sweetser2024llmVideoGames,yang2024gptGames,yang2025gptGames,maleki2024pcg}. \emph{Fa{\c{c}}ade} was retained as a historical precursor for conceptual comparison, but was not counted in the main post-LLM corpus because its architecture relied on symbolic AI, authored drama management, and pattern-based natural language processing rather than contemporary generative foundation models~\cite{mateas2003facade,mateas2002abl}. Second, because many AI-native games first appear as commercial games, independent games, demos, or platform pages rather than as academic papers, we searched public artifact sources including search engines, Steam, itch.io, official websites, developer disclosures, project pages, demo pages, industry-facing materials, and platform tags. Search terms combined AI-native and generative-AI game terms with runtime-mechanic terms and game-form terms, including ``AI game,'' ``LLM game,'' ``AI NPC,'' ``AI game master,'' ``runtime generation,'' ``interactive narrative,'' ``semantic action,'' ``procedural generation,'' ``simulation,'' ``RPG,'' ``puzzle,'' and ``sandbox.''

This process produced 98 candidate artifacts. We then conducted a second manual pass at the artifact level. When possible, we inspected or interacted with available builds, demos, web versions, project pages, videos, papers, or documentation. During this pass, we recorded how runtime AI entered the playable loop, what evidence supported the judgment, and where the case sat near the boundary of the category. Candidates were excluded when they did not provide a verifiable game experience, remained only promotional concepts, used AI only for production or static assets, offered optional chat without core-loop dependence, or otherwise failed to meet the inclusion criteria. The retained artifacts were then tagged through inductive coding and consolidated into the two axes used in Section~\ref{sec:classification}: player-facing game type (G) and dominant AI mechanic (N). The final analysis set contains 53 released, publicly playable, demo-accessible, or public research-prototype artifacts.
The public repository is maintained as a broader tracking list of AI-native and near-boundary games. It also includes unreleased titles that appear AI-native from public descriptions, but these entries are not counted in the 53-artifact corpus analyzed in this paper.

\section{Definition and Current Landscape}
\label{sec:definition}

\begin{table*}[t]
\centering
\caption{Operational criteria for identifying AI-native games.}
\label{tab:definition_criteria}
\begin{tabular}{p{0.21\linewidth}p{0.37\linewidth}p{0.34\linewidth}}
\toprule
Criterion & Operational question & Excluded cases \\
\midrule
Generative AI presence & Does the game rely on a runtime generative model rather than only traditional game AI? & Enemy AI, pathfinding, finite-state NPCs, utility AI, adaptive difficulty, and offline asset generation. \\
Core-loop dependence & Does AI output directly affect player action interpretation, content, rules, NPC behavior, world state, or outcome judgment? & Optional chat NPCs, cosmetic generation, recommendation systems, and background assistance tools. \\
Non-substitutability & Could finite authored content or deterministic rules provide the same core experience without changing the form of play? & Games where AI merely increases content volume, variation, or convenience while the underlying gameplay remains intact. \\
\bottomrule
\end{tabular}
\end{table*}

\begin{figure*}[t]
    \centering
    \includegraphics[width=1\linewidth]{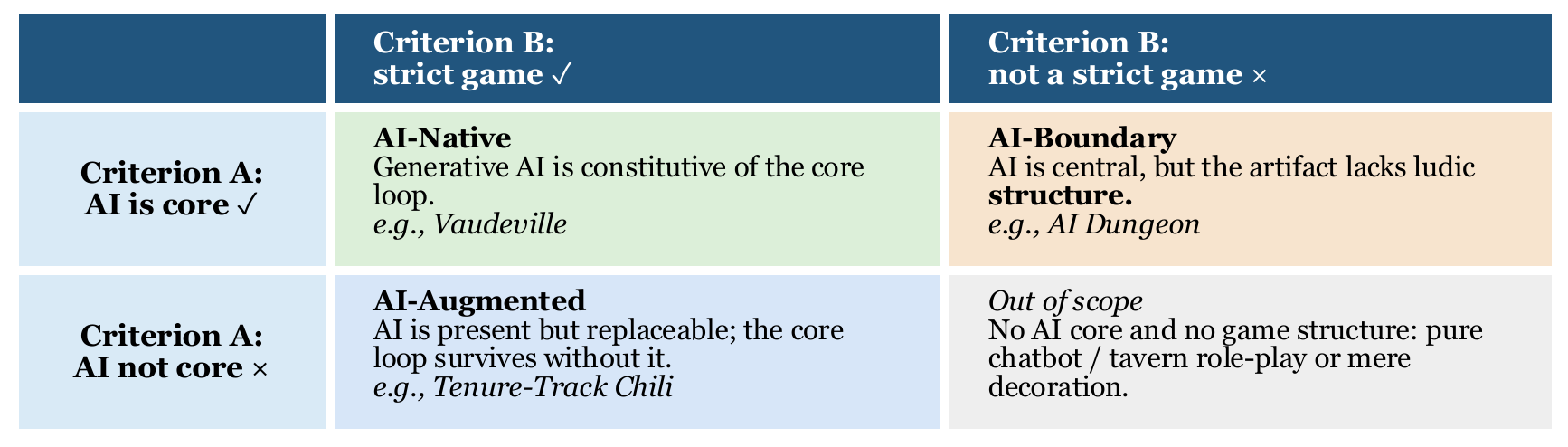}
    \caption{AI-native eligibility as a 2$\times$2 boundary matrix for candidate artifacts.
Criterion~A asks whether runtime generative AI is constitutive of the core playable activity,
rather than merely present or replaceable; Criterion~B asks whether the artifact has sufficient
ludic structure to count as a strict game.}
    \label{fig:tiers}
\end{figure*}

\subsection{Working Definition and Necessary Conditions}
We define an \emph{AI-native game} as a game in which generative AI functions as a constitutive core mechanism, such that removing the AI component would make the core gameplay impossible or would fundamentally transform its nature \cite{sun2023languageReality,gallotta2024llmGames}. This definition is intentionally narrower than the broad expression ``games using AI.'' It is not satisfied by any use of artificial intelligence in production, operation, balancing, or non-player character control. Instead, it identifies games whose core playable structure depends on runtime generative interpretation, production, or adjudication \cite{cook2016pcgPatterns,yang2024gptGames}.

The definition entails three necessary conditions. First, \emph{generative AI must be present}. The relevant system is not conventional enemy AI, pathfinding, finite-state behavior, utility-based decision-making, or adaptive difficulty control \cite{millington2009artificial,yannakakis2018aigames}. Second, \emph{the core loop must depend on AI}. Generative AI must directly participate in interpreting player actions, producing playable content or consequences, adjudicating semantic rules, controlling essential NPC behavior, or evolving the game world's state during play \cite{gallotta2024llmGames,sweetser2024llmVideoGames}. Third, \emph{the AI component must not be trivially substitutable}. A finite set of human-authored assets or a traditional deterministic mechanism should not be able to preserve the same core experience without changing the game's basic form of play.

This three-part test makes the definition both restrictive and practical. A game may contain AI-generated art, dialogue, or quests without being AI-native if these outputs are static, optional, or replaceable by conventional authored content \cite{shaker2016pcg,maleki2024pcg}. Conversely, a minimal text-based artifact may be AI-native if its primary loop depends on generative interpretation and consequential response \cite{sun2023languageReality}. The core issue is therefore not the quantity of generated content, but whether generative AI is structurally indispensable to play.

\subsection{Boundary with Adjacent Forms}

Applying these three necessary conditions will divide AI game candidate outcomes into three categories and filter out two categories that are similar but do not meet these conditions. This tiering also follows from the game-design view that playable artifacts require structured goals, feedback, mechanics, and consequential interaction, not merely open-ended content production \cite{hunicke2004mda,sweetser2005gameflow}.

\begin{itemize}
\item \textbf{AI-Native Game:}  Generative AI is a \emph{constitutive core mechanism}: the artifact satisfies all three conditions, and removing the AI makes the core loop impossible or fundamentally different in kind.
\item \textbf{AI-Augmented Game}: Runtime generative AI is \emph{present but not
constitutive}; it enriches a non-core layer while the core loop survives if the model is replaced by finite authored assets or deterministic rules. Such artifacts fail the core-loop-dependence and non-substitutability conditions.
\item \textbf{AI-Boundary Game:}  Generative AI is \emph{central to the activity, but the artifact lacks sufficient ludic structure}---goals, rules, win/lose, or state progression---to count as a game in the strict sense; these cases sit at the frontier with chatbot and tavern-style role-play.
\end{itemize}

Figure~\ref{fig:tiers} gives representative examples of the resulting tiers. 
\emph{AI Dungeon}\cite{aiDungeon2019} is placed in the AI-boundary category because generative AI is central to the activity, but the artifact remains closer to open-ended story continuation than to a strict game with stable goals, rules, win/lose conditions, and state progression. 
Conversely, \emph{Tenure-Track Chili}\cite{tenureTrackChili2026} is treated as AI-augmented: its generative layer may enrich dialogue or event variation, but the underlying play loop remains recoverable through authored content or deterministic systems. 
These cases illustrate that the classification depends not on whether AI is visible, but on whether it is both ludically structured and non-substitutable in the core loop.

This definition separates AI-native games from several adjacent categories. Traditional game AI remains largely designer-bounded: it executes authored behaviors, navigation policies, tactics, or difficulty adjustments inside a predefined representational space \cite{millington2009artificial,yannakakis2018aigames}. PCG can create levels, items, maps, quests, or layouts, but it usually operates within developer-specified grammars, parameter spaces, constraints, or search objectives \cite{hendrikx2013pcg,shaker2016pcg,togelius2011searchbased,maleki2024pcg}. AI-assisted production tools may accelerate development, but their outputs are typically fixed before play \cite{summerville2018proceduralcontentgenerationmachine,todd2023levelLLM,sudhakaran2023mariogpt}. In all of these cases, AI may be valuable, but it is not necessarily a constitutive gameplay mechanism.

The definition also distinguishes AI-native games from chatbots and tavern-style role-play systems. A conversational agent can generate open-ended text, but a game requires more than dialogue. It requires constraints, objectives, feedback, state persistence, and consequential interaction \cite{hunicke2004mda,sweetser2005gameflow}. AI-native gameplay emerges when semantic openness is coupled to ludic structure: player expressions are interpreted flexibly, but the system must convert those expressions into coherent, rule-bound, and consequential game states \cite{gallotta2024llmGames,sun2023languageReality}.

\subsection{Historical Trajectory}

\begin{figure*}[t]
\centering
\includegraphics[width=\textwidth]{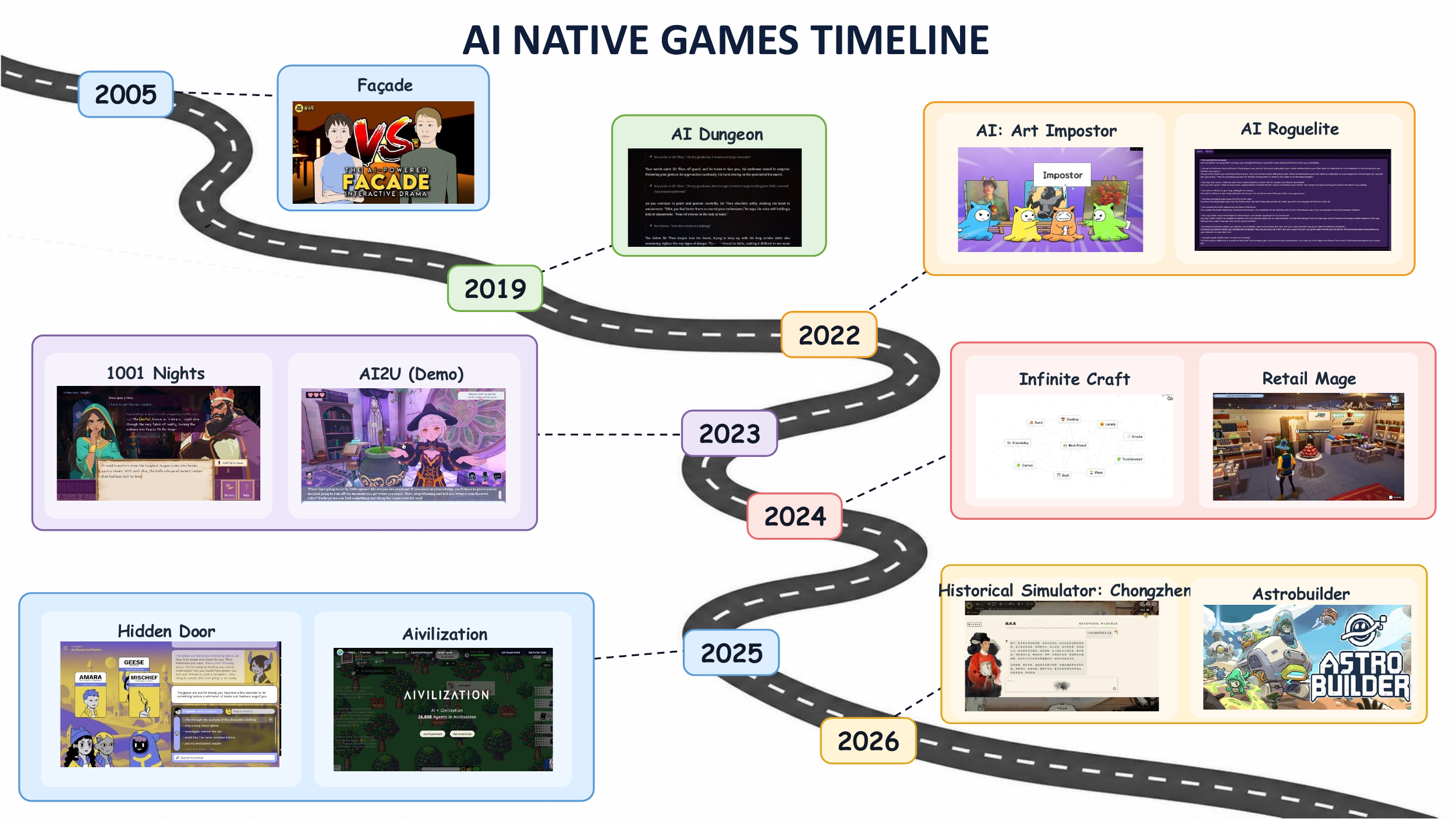}
\caption{Representative timeline of AI-native games and adjacent artifacts from early interactive drama to runtime generative AI systems.}
\label{fig:roadmap}
\end{figure*}

The history of AI-native play can be understood as a movement from authored AI behavior toward runtime semantic mediation. Early interactive drama, most notably \emph{Fa{\c{c}}ade}, demonstrated that AI could mediate social interaction and narrative tension, even though it relied on symbolic authoring, drama management, and specialized behavior architectures rather than contemporary foundation models \cite{mateas2005structuring,mateas2002abl,riedl2013interactiveNarrative}. Its significance lies less in its technical continuity with current LLM systems than in its design proposition: social language and narrative response can become the primary material of gameplay.

LLMs changed the practical conditions for AI-native design by making open-ended language input, runtime continuation, and flexible semantic response more accessible to game systems \cite{gallotta2024llmGames,yang2024gptGames}. Systems such as \emph{AI Dungeon} made open-ended text adventure widely visible in 2019 \cite{aiDungeon2019}. Later artifacts extended runtime generation into persuasion games, interrogation games, semantic crafting systems, AI game masters, educational prototypes, and multi-agent world simulations \cite{sun2023languageReality,infiniteCraft2024,park2023generativeagentsinteractivesimulacra}. Fig.~\ref{fig:roadmap} presents a timeline of representative artifacts from \emph{Fa{\c{c}}ade} to recent commercial and research prototypes. The figure should not be interpreted as an exhaustive history; rather, it highlights the shift from conversational agents toward broader world-scale and mechanics-level AI systems.

\subsection{Current Landscape and Trends}
The current landscape shows three broad trends. First, AI-native games are moving from pure text continuation toward task-oriented gameplay. Early text-adventure systems relied primarily on AI as a storyteller, while newer works use AI to support detective interrogation, social persuasion, negotiation, semantic crafting, spell interpretation, prompt-injection challenges, and AI game-master functions \cite{gallotta2024llmGames,yang2025gptGames}. This shift matters because it moves AI from content generation toward action interpretation and outcome judgment.

Second, AI-native games are increasingly hybrid. Few successful artifacts rely on unrestricted generation alone. Instead, they combine generative models with authored goals, hidden state, scenario constraints, safety filters, cached outputs, game-engine validation, or manually designed progression \cite{sun2023languageReality,gallotta2024llmGames,maleki2024pcg}. This hybrid pattern suggests that AI-native design is not a replacement of game design by models. It is a redistribution of design work toward runtime interfaces, constraints, validators, memory structures, and fallback behaviors.

Third, the field remains uneven. Many artifacts demonstrate striking interaction possibilities but still struggle with consistency, latency, evaluation, moderation, cost, and replayable depth \cite{gallotta2024llmGames,yang2025gptGames,sweetser2024llmVideoGames}. The most common current pattern is not yet a fully autonomous world simulator, but a constrained gameplay loop in which AI performs a specific role: answering as a suspect, judging a prompt, continuing a story, simulating a character, generating a playable artifact, or interpreting a player-defined action. This observation motivates the classification in the next section.

\section{Classification of AI-Native Games}
\label{sec:classification}



\begin{table*}[!t]
\centering\scriptsize
\caption{Dated AI-native games (n=53), ordered by first public availability year. The year refers to the earliest verifiable public release, Early Access launch, demo/playtest, web availability, or public research-prototype publication. Game types follow axis G (Table~\ref{tab:axis_g}); AI mechanics follow axis N (Table~\ref{tab:axis_n}).}
\label{tab:ai_native_games}
\begin{tabular}{p{0.30\linewidth}p{0.06\linewidth}p{0.15\linewidth}p{0.25\linewidth}p{0.15\linewidth}}
\toprule
Game & Year & Game type (G) & AI mechanic (N) & Status \\
\midrule
AI: Art Impostor & 2022 & G7 Party/social & N6 Generative construction & Released \\
1001 Nights & 2023 & G1 Narrative adv. & N2 Social influence & Research prototype \cite{sun2023languageReality} \\
Challenger's Odyssey & 2023 & G2 RPG & N3 Generative narr./AI GM & Released \\
Death by AI & 2023 & G7 Party/social & N3 Generative narr./AI GM & Released \\
Friends \& Fables & 2023 & G2 RPG & N3 Generative narr./AI GM & Released \\
Gandalf & 2023 & G3 Puzzle & N2 Social influence & Public prototype \\
More than words & 2023 & G8 Relationship/comp. & N2 Social influence & Released \\
Suck Up! & 2023 & G1 Narrative adv. & N2 Social influence & Released \\
Vaudeville & 2023 & G1 Narrative adv. & N1 Epistemic interaction & Released \\
Yandere AI Girlfriend Simulator & 2023 & G3 Puzzle & N2 Social influence & Released \\
AI Roguelite & 2023 & G2 RPG & N3 Generative narr./AI GM & Early Access \\
AI Asylum & 2024 & G1 Narrative adv. & N1 Epistemic interaction & Released \\
AI Game Master - Dungeon RPG & 2024 & G2 RPG & N3 Generative narr./AI GM & Released \\
AI Love Chat: Virtual Romance & 2024 & G8 Relationship/comp. & N2 Social influence & Early Access \\
AI Roguelite 2D & 2024 & G2 RPG & N3 Generative narr./AI GM & Early Access \\
Alchemy AI / Alchemic AI & 2024 & G6 Sandbox/craft & N4 Semantic adjudication & Released \\
DejaBoom! & 2024 & G1 Narrative adv. & N1 Epistemic interaction & Research prototype \cite{peng2024playerdrivenemergencellmdrivengame}\\
Doki Doki AI Interrogation & 2024 & G1 Narrative adv. & N1 Epistemic interaction & Released \\
DREAMIO: AI-Powered Adventures & 2024 & G1 Narrative adv. & N3 Generative narr./AI GM & Released \\
Hacc-Man & 2024 & G3 Puzzle & N2 Social influence & Research prototype \cite{haccman2024} \\
Infinite Craft & 2024 & G6 Sandbox/craft & N4 Semantic adjudication & Released \\
LLM-driven NPC Murder Mystery (VRST 2024) & 2024 & G1 Narrative adv. & N1 Epistemic interaction & Research prototype \cite{llmdrivenNPCmurdermystery}\\
OneSpellFitsAll & 2024 & G3 Puzzle & N4 Semantic adjudication & Public prototype \\
Retail Mage & 2024 & G5 Simulation & N4 Semantic adjudication & Released \\
Uncover the Smoking Gun & 2024 & G1 Narrative adv. & N1 Epistemic interaction & Released \\
Verbal Verdict & 2024 & G1 Narrative adv. & N1 Epistemic interaction & Released \\
Vojna & 2024 & G1 Narrative adv. & N3 Generative narr./AI GM & Released \\
AI Script: Infinite Text Adventures & 2025 & G1 Narrative adv. & N3 Generative narr./AI GM & Released \\
AI2U: With You 'Til The End & 2025 & G3 Puzzle & N2 Social influence & Early Access \\
Aivilization & 2025 & G5 Simulation & N5 Multi-agent sim. & Released \\
Caiyan (AI guessing-game hub) & 2025 & G3 Puzzle & N1 Epistemic interaction & Released \\
Civil Purgatory & 2025 & G1 Narrative adv. & N1 Epistemic interaction & Released \\
Couch Detective & 2025 & G3 Puzzle & N1 Epistemic interaction & Released \\
Hidden Door & 2025 & G1 Narrative adv. & N3 Generative narr./AI GM & Early Access \\
Minecraft Murder Mystery with LLM-driven NPCs & 2025 & G1 Narrative adv. & N1 Epistemic interaction & Public prototype\\
Pick Me Pick Me & 2025 & G7 Party/social & N2 Social influence & Released \\
Skaldsong & 2025 & G2 RPG & N3 Generative narr./AI GM & Released \\
The Last Reunion & 2025 & G1 Narrative adv. & N1 Epistemic interaction & Released \\
The Occult Detective & 2025 & G1 Narrative adv. & N1 Epistemic interaction & Released \\
Whispers from the Star & 2025 & G1 Narrative adv. & N3 Generative narr./AI GM & Released \\
AI Society & 2026 & G5 Simulation & N5 Multi-agent sim. & Released \\
Astrobuilder & 2026 & G2 RPG & N5 Multi-agent sim. & Early Access \\
Dev\_Null's Tower & 2026 & G9 Experimental & N4 Semantic adjudication & Early Access \\
Historical Simulator: Chongzhen & 2026 & G4 Strategy/mgmt & N5 Multi-agent sim. & Released \\
Hostage Down & 2026 & G1 Narrative adv. & N2 Social influence & Released \\
One Way Mirror: AI & 2026 & G1 Narrative adv. & N1 Epistemic interaction & Early Access \\
Prison Queen & 2026 & G9 Experimental & N4 Semantic adjudication & Demo/Playtest \\
Reversal Detective & 2026 & G1 Narrative adv. & N1 Epistemic interaction & Released \\
RolemIAster & 2026 & G2 RPG & N3 Generative narr./AI GM & Early Access \\
Saga \& Seeker & 2026 & G1 Narrative adv. & N3 Generative narr./AI GM & Released \\
Simulation Simulator & 2026 & G1 Narrative adv. & N2 Social influence & Demo/Playtest \\
ZeroOne Terminal & 2026 & G5 Simulation & N1 Epistemic interaction & Released \\
ZeroPrompt & 2026 & G1 Narrative adv. & N1 Epistemic interaction & Demo/Playtest \\
\bottomrule
\end{tabular}
\end{table*}

\subsection{Coding Logic and Analytic Axes}
\label{sec:coding}

\begin{table*}[!t]
\centering\small
\caption{Axis one: induced taxonomy by game type (G). The axis adopts conventional
game genres and deliberately treats themes such as ``AI detective'' as subject
matter rather than type. Counts are over the 53 dated artifacts
(Table~\ref{tab:ai_native_games}).}
\label{tab:axis_g}
\begin{tabular}{p{0.17\linewidth}p{0.08\linewidth}p{0.28\linewidth}p{0.21\linewidth}p{0.12\linewidth}}
\toprule
Game type (G) & Count & Coding criterion & Typical AI-native dependency & Example \\
\midrule
G1 Narrative Adventure & 24 (45.3\%) & The core loop advances a branching story or case through free-form text exchanges with characters or a narrator. & AI-generated dialogue, testimony, and scene continuation. & \emph{Uncover the Smoking Gun}\cite{uncoverSmokingGun2024} \\
G2 RPG & 8 (15.1\%) & The player assumes a character and advances quests, conflict, or party interactions through role-played actions. & AI-generated quests, NPC reactions, and narrative-state updates. & \emph{Friends \& Fables}\cite{friendsFables2023} \\
G3 Puzzle & 7 (13.2\%) & The core loop is solving discrete language-based challenges---guessing, combining, decoding, or bypassing a constraint. & AI evaluation of open-ended guesses, prompts, or solutions. & \emph{Gandalf} \cite{gandalf2023}\\
G4 Strategy / Management & 1 (1.9\%) & The player manages factions, resources, diplomacy, or institutions toward explicit objectives. & AI-generated diplomatic or strategic responses and event outcomes. & \emph{Historical Sim.: Chongzhen}\cite{historicalsimulatorchongzhen2026} \\
G5 Simulation & 4 (7.5\%) & Open-ended modeling of a social, historical, or everyday-life situation. & AI-generated agents, routines, and situated events. & \emph{Aivilization}\cite{aivilization2025} \\
G6 Sandbox / Creation & 2 (3.8\%) & The player freely combines or creates artifacts with no fixed win condition. & AI generation or validation of new combinations and objects. & \emph{Infinite Craft}\cite{infiniteCraft2024} \\
G7 Social Deduction / Party & 3 (5.7\%) & Multiplayer social formats built on deduction, voting, or comedic competition. & AI-generated content, role-play, and outcome arbitration. & \emph{AI: Art Impostor}\cite{artImpostor2022} \\
G8 Relationship / Companion & 2 (3.8\%) & The core loop centers on building and sustaining a long-term relationship with an AI character. & AI affective state, memory, persona consistency, and affection/relationship judgment. & \emph{More than words} \cite{morethanwords2023}\\
G9 Hybrid / Experimental & 2 (3.8\%) & Artifacts that mix genres or probe novel forms not mapping to a single conventional type. & Varies; runtime AI materially affects the playable loop. & \emph{Dev\_Null's Tower}\cite{devNullsTower2026} \\
\bottomrule
\end{tabular}
\end{table*}

The main risk in classifying AI-native games is to mistake theme for mechanic. Existing game-genre systems often collapse independent dimensions---mechanics, player structure, and subject matter---into a single label, and are therefore conceptually unclear~\cite{clarke2017genres} . This
problem is especially pronounced for AI-native games. The label ``AI detective game,'' for instance, conflates two levels: detection is the subject matter, whereas what makes the game AI-native is the mechanic by which the player must elicit hidden information from the model through open-ended questioning. Once
subject matter is mistaken for mechanic, the classification is distorted from the outset.

For this reason, we do not assume a predefined genre table but adopt an inductive coding approach~\cite{braun2006thematic}. Each artifact is examined in terms of its core loop, player activity, dominant AI function, input/output modality, and degree of runtime dependency. These observations are then consolidated into recurring game-type and AI-mechanic categories, with borderline cases revisited where they clarify category boundaries.

Some assignments were borderline. \emph{Infinite Craft}\cite{infiniteCraft2024} was close to generative construction (N6); its core AI function, however, is judging and stabilizing concept combinations, so we code it as semantic adjudication (N4). \emph{Suck Up!}\cite{suckUp2023} resembles a conversational adventure on the surface; its main loop is persuading and deceiving AI characters, so we code it as social influence (N2). These cases show that the N-axis captures the dominant AI mechanic, while genre surface and output modality are treated as secondary evidence.

We therefore code each artifact along two axes. The first is \emph{game type} (G), denoting the player-facing form of the artifact, divided into nine classes (G1 Narrative Adventure, G2 RPG, G3 Puzzle, G4 Strategy/Management, G5 Simulation, G6 Sandbox/Creation, G7 Social Deduction/Party, G8 Relationship/Companion, G9 Hybrid/Experimental); see Table~\ref{tab:axis_g}. The second is \emph{dominant AI mechanic} (N), denoting the computational function without which the core loop would not hold once generative AI is removed, divided into six classes (N1 Epistemic Interaction, N2 Social Influence, N3 Generative Narrative/AI GM, N4 Semantic Adjudication, N5
Multi-Agent Simulation, N6 Generative Construction); see
Table~\ref{tab:axis_n}.

The two axes are deliberately separated: the first captures outward form, the second the internal function that makes a game AI-native. The same subject matter can carry different mechanics, and the same mechanic can appear across different
subject matters; only by separating the two can we identify both where designs cluster and where they remain absent. The retained corpus comprises 53 dated artifacts. Unless otherwise noted, the distributional statistics below are based on these 53 entries, and the full list is given in Table~\ref{tab:ai_native_games}.

Table~\ref{tab:ai_native_games} lists all dated artifacts together with their G/N codes and release status, spanning 2022 to 2026. Such artifacts began to appear steadily from 2023 and are concentrated in 2025--2026. Most remain at an early stage (Early Access, research prototype, or demo), with only a fraction fully released. This indicates that AI-native games currently resemble an emerging field of experimentation rather than a mature category, which also justifies the use of inductive classification. 
The G-axis shows where semantic openness currently appears in game form; the N-axis shows how that openness is operationalized as a mechanic. Their intersection reveals which kinds of ludic constraint currently make generative AI playable.

\subsection{Axis One: Classification by Game Type}
\label{sec:axis-g}

\begin{figure}[t]
\centering
\includegraphics[width=\columnwidth]{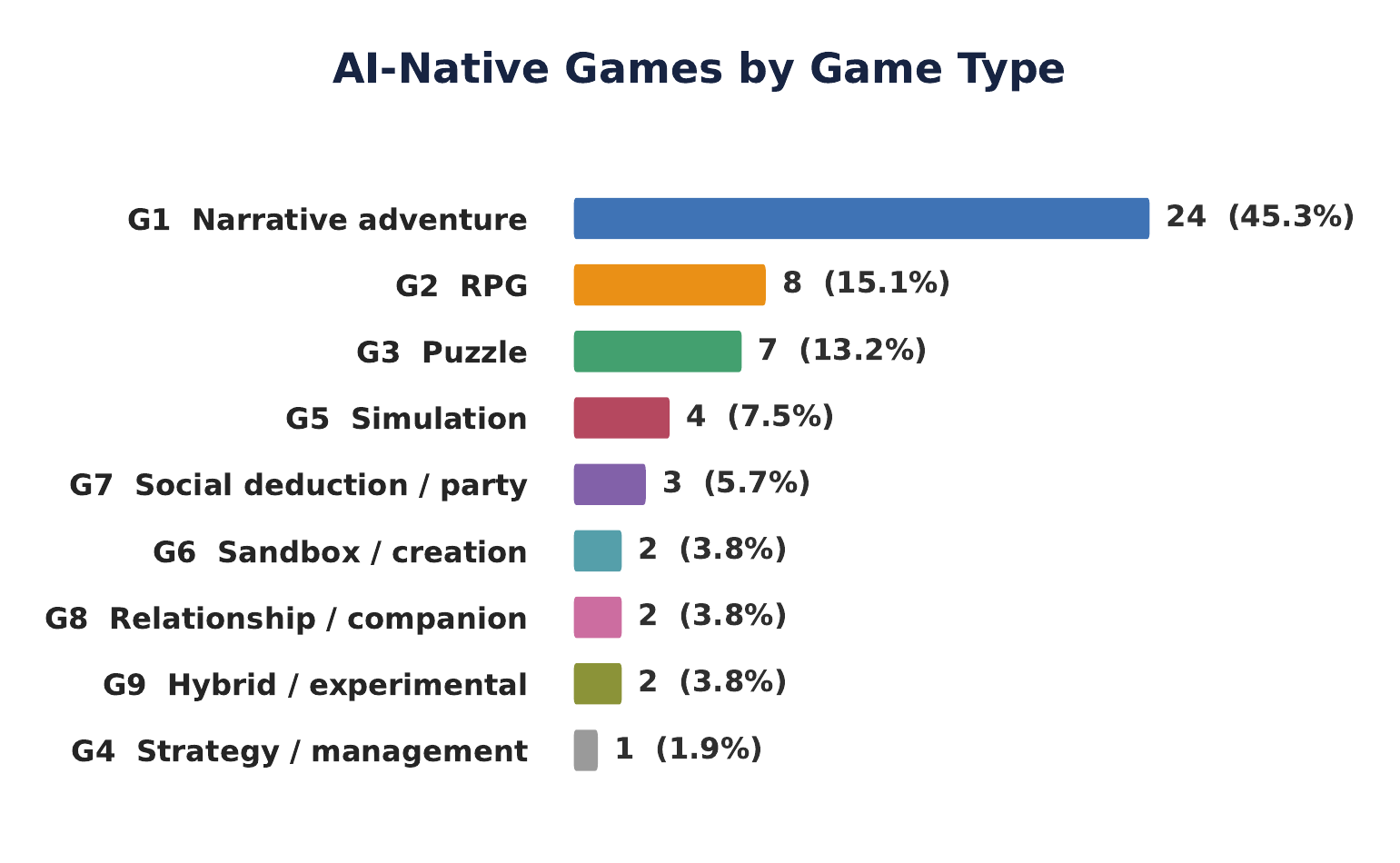}
\caption{Distribution of the dated corpus (n=53) by game type (G).}
\label{fig:game_type_distribution}
\end{figure}

Table~\ref{tab:axis_g} and Fig.~\ref{fig:game_type_distribution} show a field that remains strongly language-centered. Narrative adventure (G1) accounts for 24 of the 53 artifacts (45.3\%), followed by RPG (G2, 8, 15.1\%) and puzzle (G3, 7, 13.2\%). In these categories, natural language has a clear role in the core loop: questioning a suspect, continuing a story, role-playing an action, or judging a text-based solution. By contrast, simulation (G5, 4), social deduction/party (G7, 3), sandbox/creation (G6, 2), hybrid/experimental (G9, 2), relationship/companion (G8, 2), and strategy/management (G4, 1) form a much thinner long tail.

Current generative AI is strongest when the interaction itself is linguistic~\cite{gallotta2024llmGames,yang2025gptGames}. Yet model capability alone does not explain this distribution; game forms also differ in how well they can absorb and constrain generative uncertainty. Narrative and puzzle formats can turn ambiguity into testimony, surprise, or challenge. Strategy, simulation, sandbox construction, and long-term relationship play require more durable state, steadier balance, clearer progression, and stronger trust. Relationship/companion (G8) is especially delicate: AI companionship is already a broader product category outside games, while many companion-like systems do not provide the goals, feedback, progression, or consequential state change required here. Its small count may therefore indicate a design gap, but it also reflects the unstable boundary of the category itself.

\subsection{Axis Two: Classification by Dominant AI Mechanic}
\label{sec:axis-n}

\begin{table*}[t]
\centering\small
\caption{Axis two: induced taxonomy by dominant AI mechanic (N), i.e., the runtime AI function through which player actions are interpreted, transformed, or made consequential. Counts are over the 53 dated artifacts (Table~\ref{tab:ai_native_games}).}
\label{tab:axis_n}
\begin{tabular}{p{0.17\linewidth}p{0.08\linewidth}p{0.29\linewidth}p{0.20\linewidth}p{0.12\linewidth}}
\toprule
AI mechanic (N) & Count & Mechanic definition & Representative design pattern & Example \\
\midrule
N1 Epistemic Interaction & 17 (32.1\%) & Information seeking, investigation, interrogation, and lie detection. & Detective/interrogation loops, suspect questioning, clue elicitation. & \emph{Vaudeville}\cite{vaudeville2023}\\
N2 Social Influence & 11 (20.8\%) & Persuading, deceiving, negotiating with, pleasing, threatening, or manipulating AI characters. & Persuasion, negotiation, companion interaction, escape-by-dialogue. & \emph{Suck Up!}\cite{suckUp2023} \\
N3 Generative Narrative / AI GM & 14 (26.4\%) & Open-ended input drives plot continuation, character relations, and the story world. & Text adventures, AI TTRPG hosts, narrative RPG systems. & \emph{Hidden Door}\cite{hiddenDoor2025} \\
N4 Semantic Action \& Adjudication & 6 (11.3\%) & The AI interprets open actions, translating language into world actions, rule outcomes, or win/lose rulings. & Semantic crafting, spell judgment, prompt-injection challenges. & \emph{Infinite Craft}\cite{infiniteCraft2024} \\
N5 Multi-Agent Simulation & 4 (7.5\%) & The player observes, intervenes in, or manages a social system of multiple autonomous agents. & AI societies, colony/civilization simulations. & \emph{Aivilization}\cite{aivilization2025} \\
N6 Generative Construction & 1 (1.9\%) & The AI generates images, text, audio, objects, or worlds, and play revolves directly around the generated result. & Prompt-to-image party play, language-as-reality systems. & \emph{AI: Art Impostor}\cite{artImpostor2022} \\
\bottomrule
\end{tabular}
\end{table*}

\begin{figure}[t]
\centering
\includegraphics[width=\columnwidth]{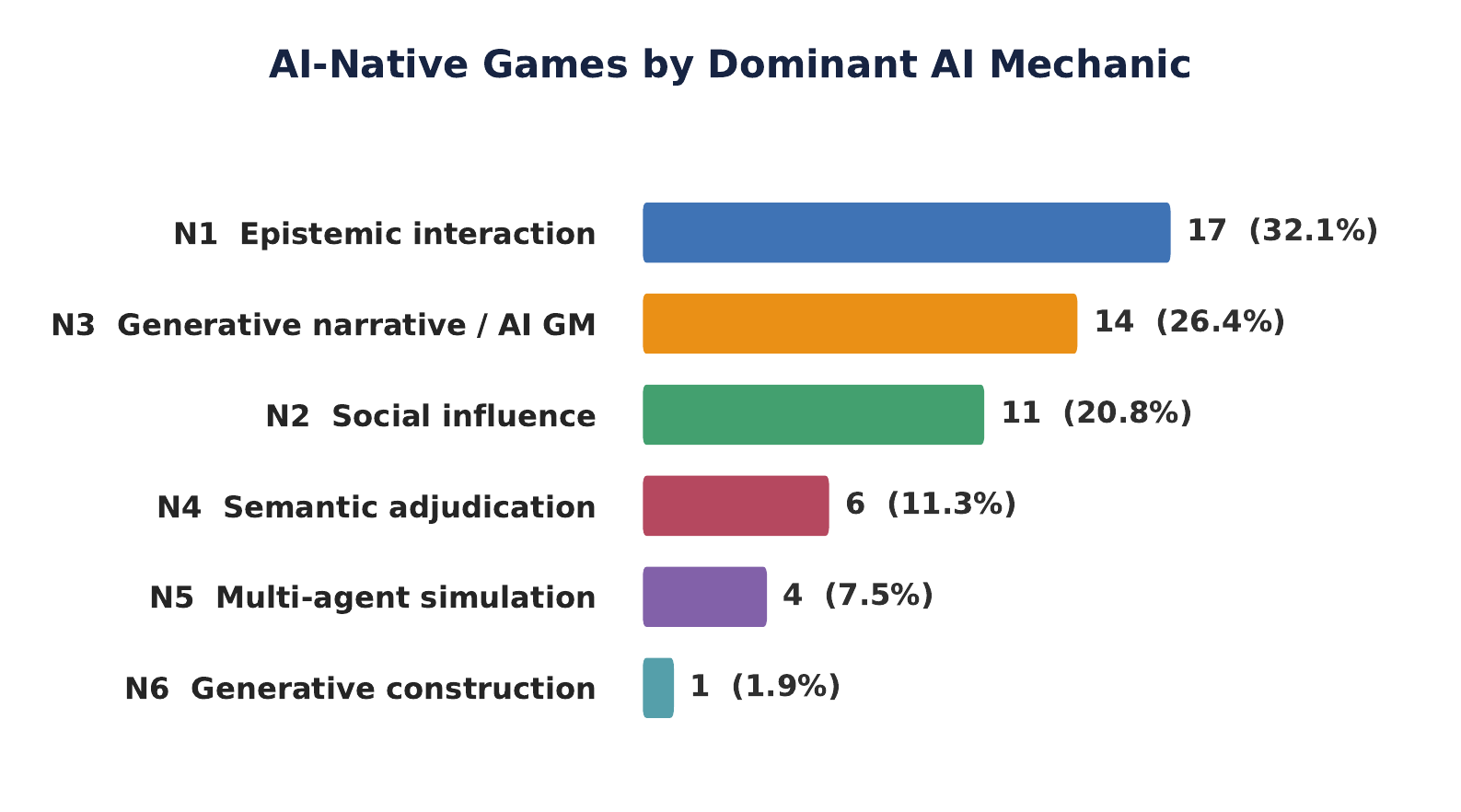}
\caption{Distribution of the dated corpus (n=53) by dominant AI mechanic (N).}
\label{fig:genai_mechanic_distribution}
\end{figure}

Axis two asks where runtime AI enters the core loop. Table~\ref{tab:axis_n} defines six dominant AI mechanics, and Fig.~\ref{fig:genai_mechanic_distribution} reports their distribution. This axis matters because AI-native play is not defined by how much content the model generates, but by how model output becomes action, feedback, state change, or judgment.

The largest groups remain close to language-based interaction. Epistemic interaction (N1, 17, 32.1\%) and generative narrative/AI GM (N3, 14, 26.4\%) together account for roughly 60\% of the corpus: one uses language to elicit hidden information, while the other uses it to continue a story world. Social influence (N2, 11, 20.8\%) is also language-centered, but makes persuasion, negotiation, or affective response the player’s task.
The smaller groups expose the harder design problem. Semantic action and adjudication (N4, 6, 11.3\%) brings AI closer to the rule layer, because the model decides whether an open-ended action succeeds, fails, combines, or changes state. Multi-agent simulation (N5, 4) and generative construction (N6, 1) remain rare because they require persistent worlds, generated artifacts, or long-running systems stable enough to support strategy.

\subsection{Cross-Axis Analysis and Discussion}
\label{sec:cross}

\begin{figure*}[t]
\centering
\includegraphics[width=\textwidth]{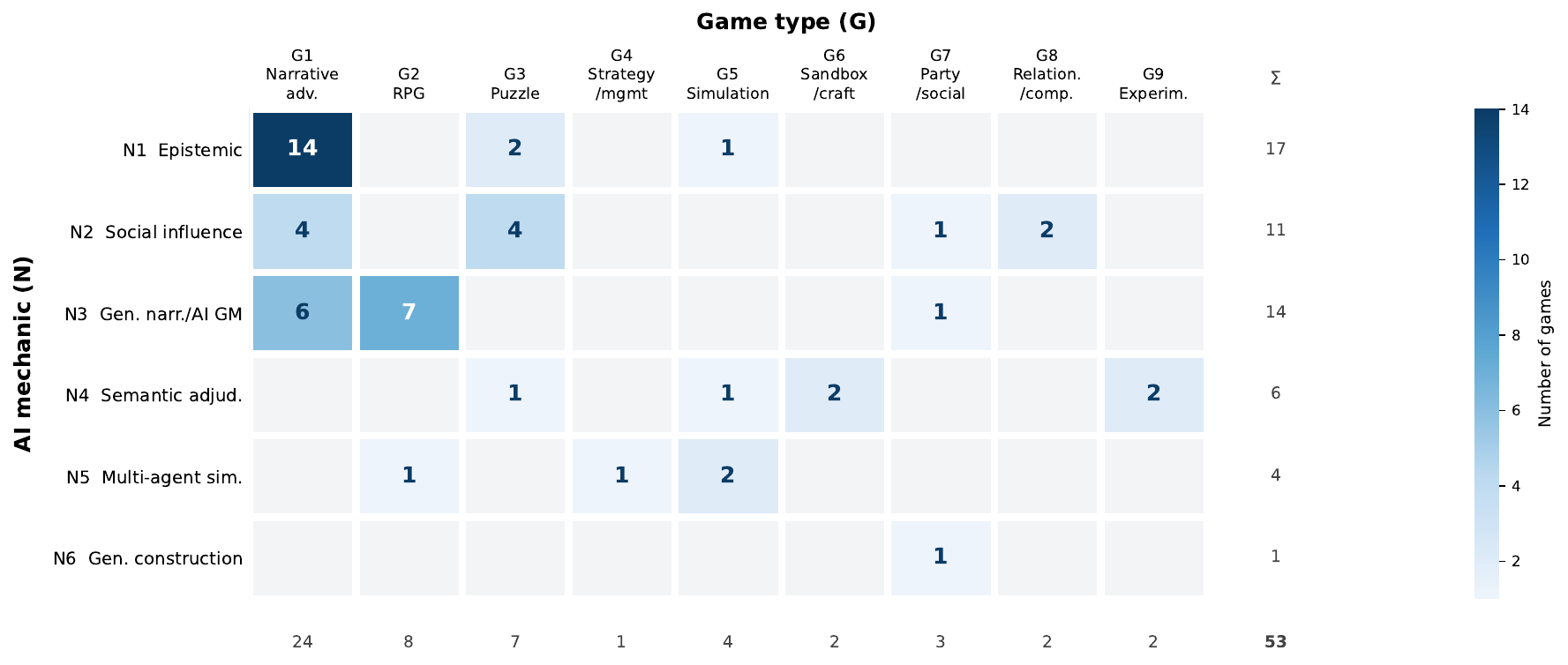}
\caption{Cross matrix of game type (G, columns) and dominant AI mechanic (N, rows)
over the 53 dated artifacts. Cell values are game counts; row/column $\Sigma$ give
marginals. Dense cells (G1$\times$N1, G1$\times$N3, G2$\times$N3) mark established
design patterns, while empty regions mark underexplored combinations.}
\label{fig:gn_heatmap}
\end{figure*}

Fig.~\ref{fig:gn_heatmap} crosses the outward form of play (G) with the AI function that enters the core loop (N). It therefore shows more than the distribution of cases: it shows where open-ended model output has so far found a playable structure. The dense cells---G1$\times$N1, G1$\times$N3, and G2$\times$N3---all sit in forms that already give language a clear role in play. Investigation games can bind model responses to suspects, clues, and hidden information; narrative adventures can bind continuation to characters, scenes, and progression; RPG-like systems can bind generated events to roles, quests, and world state. By contrast, the sparse regions mark places where openness is harder to hold in place. Relationship/companion play needs memory, trust, and progression without becoming only open-ended chat; generative construction needs generated artifacts to become objects of play rather than one-off outputs; multi-agent simulation needs model-driven entities to stay coherent enough for players to read, intervene, and plan.

Taken together, the matrix suggests that the central problem of AI-native design is not how much a model can generate, but where generated meaning can be made playable. Current artifacts work best when semantic openness is tied to recognizable game structures: suspects and clues, quests and roles, puzzles and judgments, states and consequences. AI-native game design is therefore the design of semantic openness under ludic constraint.

\section{Challenges and Future Roadmap}
\label{sec:roadmap}



Generative AI can readily produce content and responses, but such generation does not automatically become stable gameplay. Section~\ref{sec:roadmap} therefore organizes the roadmap around two connected layers: the design architecture of AI-native games and the runtime conditions that make such designs viable. On the design side, Section~\ref{subsec:controllable} examines the tension between open-ended generation and game rules; Section~\ref{subsec:mechanics} asks how AI can move from content generation into mechanics; Section~\ref{subsec:modality} extends the discussion beyond text-based interaction; and Section~\ref{subsec:multiagent} scales the problem from single-character interaction to multi-agent worlds. On the runtime side, Section~\ref{subsec:economics} discusses inference latency, marginal cost, deployment, and business models, while Section~\ref{subsec:repro} considers model dependence, portability, and reproducibility. Ethical and safety concerns are treated separately in Section~\ref{sec:ethics}.


\subsection{Rule-Bounded Controllable Generation}
\label{subsec:controllable}


The core technical challenge is the tension between generative openness and game-state reliability. Generative AI allows players to express actions flexibly, but games still depend on rules, state, and consistency. Which rules must remain fixed? Where should players be given semantic freedom? What should happen when a model invents an event, item, quest, or rule outcome that the current game state cannot support? These are not only engineering questions, but also design questions.


One possible direction is a generate-and-verify pipeline: the model first proposes an event, quest, action interpretation, or rule outcome, and the game system then checks whether it is compatible with the current rules and state. For example, does the required object exist? Can the player reach the location? Does the outcome contradict prior narrative or world state? If validation fails, the system may ask the model to revise the proposal, ask the player to clarify the action, or fall back to authored content~\cite{madaan2023selfrefineiterativerefinementselffeedback}. Existing techniques offer partial support: structured generation and function calling make model outputs easier for game engines to parse and execute; constraint solving and simulation can test whether player actions or generated scenarios are valid; retrieval, fine-tuning, and preference learning help keep generation grounded in prior events and genre expectations~\cite{willard2023efficientguidedgenerationlarge,geng2024grammarconstraineddecodingstructurednlp,schick2023toolformerlanguagemodelsteach,yao2023reactsynergizingreasoningacting,smith2011asp,togelius2011searchbased,lewis2021retrievalaugmentedgenerationknowledgeintensivenlp,ouyang2022traininglanguagemodelsfollow,rafailov2024directpreferenceoptimizationlanguage}. \emph{Infinite Craft}~\cite{infiniteCraft2024} offers a lightweight example: by caching semantic combinations, it makes once-generated results stable for later play. The design problem is to decide what the model may invent, what the system must verify, and which rules and states must remain fixed.

The same tension also appears over longer time scales as memory instability: even if a generated action is locally valid, it may later fail to persist as part of the game state. In ordinary chatbots, forgetting mainly weakens conversational continuity; in AI-native games, it can break state persistence and causal coherence. If a model forgets prior decrees, relationships, resource changes, or long-term plans, player actions no longer accumulate into a reliable game state. The player must then repeatedly remind the system of its own history, which turns strategy into prompt maintenance and undermines agency \cite{zhong2023memorybankenhancinglargelanguage,packer2024memgptllmsoperatingsystems}. 

\subsection{Mechanics-Level Innovation}
\label{subsec:mechanics}

For AI-native games to move beyond content generation and offer genuinely new forms of playability, the key question is mechanics-level innovation: how can AI become part of the rules of play? Many AI games still use AI at the content layer, generating dialogue, quest descriptions, character reactions, or narrative fragments. Such uses can increase variation, but they do not always change what the player is actually playing. A more promising direction is to involve AI in action interpretation, rule adjudication, state expansion, and feedback generation, so that open-ended player input can directly change the game system.

Semantic action and adjudication (N4) illustrates this possibility. \emph{OneSpellFitsAll}\cite{oneSpellFitsAll2024} and \emph{Prison Queen}\cite{prisonQueen2026} translate player-authored language into puzzle progress or combat damage. \emph{Infinite Craft}\cite{infiniteCraft2024} turns concept combinations into reusable new elements, while \emph{Dev\_Null's Tower}\cite{devNullsTower2026} maps player bug reports into level modifications. Across these cases, AI takes on a mechanical function between player input and game consequence, determining how open expression enters state, feedback, progression, and strategic loops.

The difficulty is that open input must be organized into learnable play. Players may express themselves freely, yet they still need to understand how the system responds. When AI only produces one-off surprises, it remains a source of novelty rather than a durable mechanic. Mechanics-level AI innovation should make language, concept combinations, and semantic actions enter rules and state in a stable enough way for players to test, learn, plan, and develop strategies around the system's mode of interpretation.

\subsection{Beyond Text Modality}
\label{subsec:modality}


One immediate challenge is the dominance of a single modality, which may also weaken playability. The artifacts surveyed in this paper are overwhelmingly text-driven and often rely on typed or spoken input from the player. For players accustomed to conventional game interaction, this mode can feel less natural and may interrupt the rhythm of feedback. In some artifacts, generative AI appears as a surface-level addition that never becomes structurally embedded in the playable loop, limiting its role to novelty instead of mechanic design.

First, AI-native games should explore forms of interaction beyond text, including speech~\cite{radford2022robustspeechrecognitionlargescale}, images~\cite{rombach2022highresolutionimagesynthesislatent}, video, embodied world models, and interface-level actions. Existing examples already point in this direction: \emph{Whispers from the Star}~\cite{whispersStar2025} uses real-time voice conversation as a central interaction mode, while \emph{1001 Nights}~\cite{sun2023languageReality} combines player-authored storytelling with generated visual and narrative consequences. Second, designers should develop mechanics in which modality and play are naturally integrated. For example, \emph{AI: Art Impostor}~\cite{artImpostor2022} connects prompt-to-image generation with drawing, guessing, deception, and social judgment. More broadly, generative AI can be coupled with interface design, spatial action, rule feedback, and state changes, rather than reducing AI-native play to repeated text entry.

\subsection{Multi-Agent Worlds and Dynamic Balancing}
\label{subsec:multiagent}

AI-native games can also use generative models to support multi-agent worlds and adaptive pacing. Instead of treating NPCs as isolated dialogue partners, a game may give them goals, memories, social ties, and local routines, allowing towns, factions, economies, and companions to change over time \cite{park2023generativeagentsinteractivesimulacra,wang2023voyageropenendedembodiedagent}. Memory-based agent architectures are especially relevant for companions and persistent social worlds, where past interactions need to remain available across later play \cite{packer2024memgptllmsoperatingsystems}. Since each additional agent increases inference cost and creates more opportunities for inconsistency or unsafe behavior, practical systems will likely need layered simulation: lightweight routines for background behavior, and foundation-model inference for interactions that matter to the player or the story. Large-scale agent projects such as \emph{Aivilization}\cite{aivilization2025} and \emph{AI Society}\cite{aiSociety2026} point toward this direction. A similar logic applies to dynamic balancing. Traditional adaptive systems, such as the AI Director in \emph{Left 4 Dead}\cite{valve2009aidir}, monitor player state to adjust pacing; AI-native games can extend this idea from enemy spawns and resource pressure to semantic pacing, such as when to introduce a mystery, escalate a conflict, summarize past events, limit disruptive actions, or leave room for player improvisation. The challenge is to make these adaptations legible enough to support play without making players feel manipulated.



\subsection{Inference Optimization, Economic Viability, and Business Models}
\label{subsec:economics}

Latency is where inference cost becomes visible as a playability problem. In many dialogue-centered AI-native games, typed interaction creates a fragile turn rhythm: the player enters an utterance or action, waits for model inference, and only then receives feedback. When this delay is long or unpredictable, it interrupts real-time responsiveness, weakens presence, and makes the loop feel less like play and more like waiting for a remote service. Historical Simulator: Chongzhen \cite{historicalsimulatorchongzhen2026} partially sidesteps this problem through an edict-and-memorial structure, in which player commands are resolved through periodic state updates and reports rather than continuous conversational turn-taking. This design can buffer some inference latency, but it does not solve the harder problem faced by dialogue-centered AI-native games: current model inference is still often too slow and variable to support tightly timed feedback loops.

Generative inference has a usage-dependent marginal cost, often tied to token count, request volume, model size, and inference time. This makes economic viability both an engineering and a commercial problem. On the engineering side, a practical system should avoid calling large models for every interaction. Frequent and low-risk behavior can be handled by local small models or traditional AI, while larger models are reserved for moments where semantic uncertainty matters. Caching is also central: once a pair of elements in \emph{Infinite Craft}\cite{infiniteCraft2024} produces a result, that result remains stable for later play. The same principle can apply to rule judgments, generated entities, NPC memories, and world changes, turning temporary generation into durable game state. Other optimization techniques, including distillation \cite{hinton2015distillingknowledgeneuralnetwork}, quantized local inference \cite{frantar2023gptqaccurateposttrainingquantization,dettmers2023qloraefficientfinetuningquantized}, speculative decoding \cite{leviathan2023fastinferencetransformersspeculative}, and batching for multi-agent scenes \cite{kwon2023efficientmemorymanagementlarge}, can further reduce cost and latency. The broader goal is to reposition inference as a design problem: its latency, cost, and variability should shape mechanic design instead of being treated as a hidden backend concern.

The commercial side is less settled. Because the marginal-cost constraint applies to any game that calls a model at runtime, the evidence here spans the Native, Boundary, and Augmented tiers (for instance, \emph{AI Dungeon}\cite{aiDungeon2019} is a Boundary title). Two routes have emerged. On the cost side, developers externalize inference: local/on-device execution removes the marginal cost (e.g., \emph{Suck Up!}\cite{suckUp2023}, \emph{AI Roguelite}\cite{aiRoguelite2024}), or a bring-your-own-API-key model lets the player pay the provider directly. On the revenue side, they adopt recurring charges: freemium plus tiered subscriptions (e.g., \emph{AI Dungeon}\cite{aiDungeon2019}, the monthly \emph{Friends \& Fables}\cite{friendsFables2023}, and \emph{Skaldsong}\cite{skaldsong2025} with a free daily quota) and prepaid credits or tokens consumed per AI action, often combined into a subscription-plus-credits hybrid. Broadly, packaged titles lean toward local inference and BYOK, while platform titles lean toward subscriptions and credits. Token-based billing is not yet a mature business model, and inference cost may exceed players' willingness to pay---which makes localization and on-demand inference a necessity that is at once technical and commercial.

\subsection{Model Dependence, Portability, and Reproducibility}
\label{subsec:repro}

AI-native games depend heavily on external foundation models, which creates several risks. Different base models may materially change the experience; access to a specific model may be limited by region, availability, or compliance constraints; and closed hosted models may change behavior under the same name \cite{chen2023chatgptsbehaviorchangingtime}. These dependencies make AI-native games hard to test, reproduce, and preserve. Evaluation should therefore be grounded in the player-facing consequences of these conditions: whether the game can still sustain coherent, responsive, and enjoyable play
\cite{liang2023holisticevaluationlanguagemodels}. The most important criteria are rule consistency, response latency, and long-session memory stability: does the game apply its rules reliably, respond quickly enough to sustain the loop, and remember player actions over extended play? Benchmarks for AI-native games should therefore test extended play sessions, adversarial player behavior, and changes across model versions. These tests should ask whether the game still applies its rules, responds in time, and preserves player history when the model is stressed. The long-term goal is to evaluate AI-native games as playable systems, not only as model demonstrations.

\section{Ethics, Safety, and Regulatory Impacts}
\label{sec:ethics}

These risks also vary by mechanic. In epistemic interaction (N1), hallucination is especially consequential because it can enter the game as false testimony, unstable clues, or corrupted evidence. For social influence mechanics (N2), persistent memory and personalization can turn ordinary engagement design into manipulation, attachment pressure, or unclear consent. For semantic adjudication (N4), the model’s judgment directly determines the success or failure of player actions. In this setting, opacity affects not just how decisions are explained, but how fair those decisions appear to players. For multi-agent simulation (N5), risk no longer comes from a single generated response alone, but from interactions among agents, memories, goals, and player interventions. Safety evaluation for AI-native games should therefore follow the dominant AI mechanic, not only the underlying model.

\subsection{General Foundation Model Risks}
AI-native games inherit the general risks of foundation models, including bias, toxicity, hallucination, privacy leakage, and vulnerability to adversarial prompting~\cite{gallotta2024llmGames,weidinger2021ethicalsocialrisksharm}. These risks matter here because model outputs are absorbed into play as roles, states, and consequences, where they can reshape the game itself. A biased NPC can become a persistent social actor, not just a source of problematic dialogue~\cite{bender2021parrots}. A hallucinated quest may consume player time, break progression, or corrupt world state, giving the error consequences beyond factual inaccuracy~\cite{ji2023hallucination}. A privacy leak can also extend beyond a chat transcript, encompassing intimate role-play, voice, emotion, player habits, and behavioral profiles~\cite{carlini2021extractingtrainingdatalarge}.

\subsection{Real-Time Generation and Runtime Moderation}

Conventional content rating and moderation assume that most game content can be inspected before release. AI-native games weaken that assumption. Dialogue, quests, images, and even rule outcomes may be produced during play, after certification and outside the direct view of developers. Safety therefore becomes a runtime design problem rather than only a pre-release review process~\cite{inan2023llamaguardllmbasedinputoutput,markov2023holisticapproachundesiredcontent}.

Prompt injection is the clearest example of this shift~\cite{perez2022ignorepreviouspromptattack,greshake2023youvesignedforcompromising}. Players may try to make NPCs reveal hidden prompts, bypass safety filters, generate prohibited content, or alter rules~\cite{wei2023jailbrokendoesllmsafety,zou2023universaltransferableadversarialattacks}. In conventional games, exploits usually target code, data, or level geometry; in AI-native games, language context also becomes an attack surface. Hacc-Man~\cite{haccman2024} and Lakera's Gandalf~\cite{gandalf2023} show that prompt injection can even be turned into a playable educational mechanic. In commercial games, the same behavior becomes harder to contain, because the system must preserve playful interaction while preventing unsafe or platform-violating outputs.

Runtime moderation therefore needs to be part of the game architecture. Tool permissions, output classifiers, player-report pipelines, age-appropriate filters, adversarial testing, and incident logging all help, but none is sufficient alone. Safety cannot be left to the model's refusal behavior; it has to be designed into the loop of play.

\subsection{Behavioral Manipulation and Addiction Vulnerabilities}
AI-native games can personalize interaction at a depth unavailable to static games. A model can learn what a player likes, fears, avoids, purchases, and emotionally responds to. Combined with reward schedules, social bonding, and persistent NPCs, this creates risks of behavioral manipulation \cite{mathur2019darkpatterns}. A companion NPC could pressure a player to return. A dynamic economy could personalize scarcity \cite{zendle2018lootboxes}. A narrative system could adapt cliffhangers to a player's vulnerabilities. Children and adolescents are especially exposed because they may treat conversational agents as social beings \cite{laestadius2022replika}.

The ethical issue is not that personalization is inherently bad. Adaptive accessibility, player modeling, and emotionally responsive narratives can be beneficial. The issue is opacity and asymmetry. If the system optimizes engagement, spending, or retention using intimate behavioral data, players need meaningful consent, boundaries, and controls. AI-native games should provide transparency about model use, memory, data retention, and monetization. They should allow players to inspect or reset persistent memories, opt out of sensitive personalization, and distinguish fictional role-play from persuasive system behavior.

\subsection{Fairness, Explainability, and Player Trust}
Games are built on trust. Players accept failure when they believe the system is fair. AI judges complicate this trust because their decisions may be stochastic, opaque, or inconsistent \cite{zheng2023judgingllmasajudgemtbenchchatbot}. If a spell works once and fails later under similar conditions, the player needs an explanation. If an AI detective suspect changes testimony, the player must be able to distinguish deliberate lying from model drift. If a generated level is unwinnable, the system should detect and repair it.

Explainability in AI-native games should be gameplay-oriented. It does not require exposing model weights or chain-of-thought \cite{wei2023chainofthoughtpromptingelicitsreasoning}. It requires giving players actionable reasons: the door resisted because the lock is arcane; the NPC refused because trust is low; the treaty failed because food supply is insufficient; the spell backfired because the requested effect exceeds available mana. Explanations should connect model judgments to game state. This is both a design requirement and a safety requirement \cite{doshivelez2017rigorousscienceinterpretablemachine}.

\subsection{Regulatory and Preservation Challenges}

AI-native games unsettle several assumptions behind existing game governance. Age ratings, data protection, consumer disclosure, and content liability are usually assessed against a relatively stable product, but runtime generation means that dialogue, quests, images, and rule outcomes may change after release~\cite{euaiact2024}. This makes responsibility harder to assign: if a model generates infringing dialogue, unsafe imagery, or a harmful quest during play, the relevant actor may be the developer, platform, model provider, or player.

The same instability complicates copyright, authorship, and creative labor. If a model generates dialogue, images, quests, or music during play, ownership and liability become difficult to define; if a player co-authors a world through prompts, their rights are unclear~\cite{henderson2023foundationmodelsfairuse}. AI-native games also affect attribution, compensation, and training-data licensing for writers, artists, voice actors, and composers, as platform AI-disclosure rules already indicate~\cite{steam2024aidisclosure}.

Preservation raises a related but distinct problem. A game dependent on a proprietary hosted model may not remain playable in the future~\cite{newman2012bestbefore,chen2023chatgptsbehaviorchangingtime}. Even if the executable survives, its behavior may change when the model endpoint, safety policy, retrieval database, or prompt template changes. 
For AI-native games, preserving the executable is no longer enough; without the model, prompts, state structure, and traces of generated play, the software may survive while the playable experience disappears. Preservation matters because AI-native games capture a formative moment in a medium whose technical foundations, design conventions, and cultural meanings are still changing rapidly.

\section{Conclusion}
\label{sec:conclusion}

This survey has defined AI-native games as a gameplay category in which runtime generative AI becomes a constitutive part of the core play loop. By distinguishing AI-native games from AI-assisted production, AIGC-enhanced games, traditional PCG, optional chat NPCs, and chatbot-like role-play systems, we have clarified the central promise and difficulty of the category: AI-native games place foundation models inside the fragile structure of rules, goals, memory, pacing, fairness, and player agency.

The analyzed corpus shows a field that is growing quickly but still unevenly developed. Current artifacts cluster around language-forward forms, especially narrative adventure, RPG, puzzle play, epistemic interaction, social influence, and generative narrative. These are the places where open-ended language can most easily be tied to recognizable game structures. By contrast, semantic adjudication, multi-agent simulation, generative construction, long-term relationship play, and strategy or management remain thinner and less stable. Several challenges therefore remain central. Runtime generation is still difficult to control, especially when outputs must remain coherent with game state and rules. Open-ended semantic input often exceeds the representational capacity of existing game systems. Persistent memory, world-state tracking, and long-term consequence management remain fragile. Inference cost, latency, model portability, and deployment constraints shape which forms of AI-native play are practically viable. Safety, evaluation, and regulation also become harder when content is generated during play rather than authored before release.

Addressing these challenges will require both technical and design-level progress. Controllable generation should be paired with explicit state models, validators, and rule-aware constraints. Foundation models should be integrated with classical game AI, simulation systems, PCG pipelines, retrieval, planning, and causal reasoning, rather than treated as standalone engines. Evaluation should move beyond output quality toward playability, fairness, robustness, safety, and player experience. More broadly, designers will need methods for prototyping mechanics around semantic action, negotiated rules, persistent memory, emergent narrative, and multi-agent worlds. If these challenges can be sufficiently addressed, AI-native games may become more than a new subgenre. They may mark a broader shift in games and interactive entertainment: from fixed authored worlds, to procedurally generated worlds, to responsive worlds that can interpret, negotiate with, and evolve around player intent. In such worlds, AI would not merely generate more content; it would help create forms of play that are more adaptive, expressive, personal, and alive. The long-term potential of AI-native games is therefore not only to extend the boundaries of existing game design, but to open a new frontier for how digital worlds are built, inhabited, and experienced.

\bibliographystyle{IEEEtran}
\bibliography{bibtex/bib/IEEEabrv,bibtex/bib/references}

\end{document}